\documentclass[letterpaper]{article} % DO NOT CHANGE THIS
\usepackage[]{aaai24}  % DO NOT CHANGE THIS
\usepackage{times}  % DO NOT CHANGE THIS
\usepackage{helvet}  % DO NOT CHANGE THIS
\usepackage{courier}  % DO NOT CHANGE THIS
\usepackage[hyphens]{url}  % DO NOT CHANGE THIS
\usepackage{graphicx} % DO NOT CHANGE THIS
\usepackage{natbib}  % DO NOT CHANGE THIS AND DO NOT ADD ANY OPTIONS TO IT
\usepackage{caption} % DO NOT CHANGE THIS AND DO NOT ADD ANY OPTIONS TO IT
\usepackage{algorithm}
\usepackage{algorithmic}

\usepackage{newfloat}
\usepackage{listings}

\usepackage{amsmath} %for cases
\usepackage{amssymb} %for mathbb
\usepackage{tikz}    %for diagrams
\usepackage{subcaption}
\usetikzlibrary{automata, arrows.meta, positioning}
\usepackage{array} % for equal-cell tables
\usepackage[thinlines]{easytable}  % for equal-cell tables

\DeclareCaptionStyle{ruled}{labelfont=normalfont,labelsep=colon,strut=off} % DO NOT CHANGE THIS
\floatstyle{ruled}
\newfloat{listing}{tb}{lst}{}
\floatname{listing}{Listing}
\title{Numeric Reward Machines}
\author{
    Kristina Levina\textsuperscript{\rm 1,\rm 2},
    Nikolaos Pappas\textsuperscript{\rm 1},
    Athanasios Karapantelakis\textsuperscript{\rm 2},
    Aneta Vulgarakis Feljan\textsuperscript{\rm 2},\\
    Jendrik Seipp\textsuperscript{\rm 1}
}
\affiliations{
    \textsuperscript{\rm 1}Linköping University, Linköping, Sweden\\
    \textsuperscript{\rm 2}Ericsson Research, Stockholm, Sweden\\
    \{kristina.levina, nikolaos.pappas, jendrik.seipp\}@liu.se
     \{aneta.vulgarakis, athanasios.karapantelakis\}@ericsson.com
}

\nocopyright

\newcommand{\numericbool}{\emph{numeric--Boolean}}
\newcommand{\Numericbool}{\emph{Numeric--Boolean}}
\newcommand{\numeric}{\emph{numeric}}
\newcommand{\Numeric}{\emph{Numeric}}
\newcommand{\bool}{\emph{Boolean}}
\newcommand{\props}{\ensuremath{{\mathcal{P}}}}
\newcommand{\psa}{\ensuremath{{\props\mathit{SA}}}}

\newcommand{\inlinecite}[1]{\citet{#1}}

\newcommand{\dec}{\hspace{-.05em}\raisebox{.15ex}{\footnotesize$\downarrow$}}

\begin{document}

\maketitle

\begin{abstract}
Reward machines inform reinforcement learning agents about the reward structure of the environment and often drastically speed up the learning process.
However, reward machines only accept Boolean features such as \texttt{robot-reached-gold}.
Consequently, many inherently numeric tasks cannot profit from the guidance offered by reward machines.
To address this gap, we aim to extend reward machines with numeric features such as \texttt{distance-to-gold}.
For this, we present two types of reward machines: \numericbool{} and \numeric{}. In a \numericbool{} reward machine, \texttt{distance-to-gold} is emulated by two Boolean features \texttt{distance-to-gold-decreased} and \texttt{robot-reached-gold}. In a \numeric{} reward machine, \texttt{distance-to-gold} is used directly alongside the Boolean feature \texttt{robot-reached-gold}.
We compare our new approaches to a baseline reward machine in the Craft domain, where the numeric feature is the agent-to-target distance.
We use cross-product $Q$-learning, $Q$-learning with counter-factual experiences, and the options framework for learning.
Our experimental results show that our new approaches significantly outperform the baseline approach. Extending reward machines with numeric features opens up new possibilities of using reward machines in inherently numeric tasks.
\end{abstract}

\section{Introduction}

Reinforcement learning (RL) is the process of learning an optimal policy by interacting with an environment \cite{sutton2018reinforcement}. The agent receives rewards from the environment based on its actions. The goal is to learn a policy that maximises the expected accumulated reward over time. The reward function is crucial for the agent to learn an optimal policy. However, designing it is often challenging and time-consuming.

Several methods have been developed to address this challenge. For example, instead of defining a full reward function, one can specify requirements for the agent's behaviour in logic formulas, called \emph{specifications} \citep{jothimurugan2023specification,krasowski2023provably}. The literature describes various specification languages and approaches for compiling specifications to rewards. \citet{jothimurugan2021compositional} designed a compositional learning approach, called DIRL, for translating the supplied specifications into an abstract graph on which high-level planning is then performed for model-based RL. Furthermore, \citet{illanes2020symbolic} introduced high-level symbolic action models. Then, they used automated planning for guiding hierarchical RL (HRL) techniques. Another approach for encoding the required RL agent behaviour is to use temporal logic, for example, truncated linear temporal logic \cite{li2017reinforcement} and signal temporal logic (STL) \cite{balakrishnan2019structured}.

In the aforementioned studies, the use of specifications for guiding the RL agent has shown a remarkable performance in comparison with entirely manual reward design. In addition, techniques such as automatic reward shaping \cite{ng1999policy} are used to reshape the reward function such that an optimal policy is found faster. However, these approaches cannot handle non-Markovian rewards \cite{skalse2023limitations} that naturally arise in sparse-reward or partially observable environments \cite{kazemi2022translating}.

To handle non-Markovian rewards, automaton-based approaches have been developed for high-level guidance of the RL agent. For example, \citet{jothimurugan2019composable} designed a specification-to-reward compiler, called SPECRL, with a task monitor, which is an automaton that keeps track of completed tasks. Later, \citet{icarte2022reward} introduced reward machines (RMs) that are automata representing the environment on a high level. RMs reflect sub-goals that the agent is supposed to achieve on the way to the main goal. RL algorithms with access to an RM can simulate experiences at all RM states using only a single interaction with the environment. This can drastically speed up learning by making it more sample-efficient. RMs not only handle non-Markovian tasks but are also advantageous over plain reward functions and STL-specified reward functions \cite{unniyankal2023rmlgym}.

However, RMs accept only Boolean features and thus cannot be used in inherently numeric tasks. Our work intends to address this gap by extending RMs with numeric features. We build upon the work by \citet{icarte2022reward} and use their code for our experiments \cite{Icarte2021}. We introduce two types of RMs: \numericbool{} and \numeric{}. To illustrate them with an example, consider a numeric feature $d$ that represents the distance to the target. In a \numericbool{} RM, $d$ is emulated by two Boolean features $\dec d$ and $d$=$0$. The former signals whether $d$ decreases, and the latter signals whether the RL agent has reached the target. The agent receives a positive reward when one of these features becomes true. In a \numeric{} RM, $d$ is used directly together with the Boolean feature $d$=$0$, and the agent is rewarded with $-d$ after each action. We compare our proposed RMs with the original \emph{Boolean} RM developed by \citet{icarte2022reward}.

In summary, our main contribution is the introduction of numeric features into RMs. We show empirically using the Craft domain \cite{andreas2017modular} that \numericbool{}-RM and \numeric{}-RM-based methods outperform \bool{}-RM-based methods in terms of learning speed. RL agents can now leverage RMs in inherently numeric tasks, such as energy optimisation \cite{gao2022operational}. This expansion of the RM applicability can further promote research in RMs.

\section{Background}

We begin by providing background information on RL and RMs. For more details on the topics, please refer to \citet{sutton2018reinforcement} and \citet{icarte2022reward}, respectively.

\subsection{Reinforcement Learning}

Single-agent RL tasks are generally formalised via Markov decision processes (MDPs) \cite{sutton2018reinforcement}. An MDP is a tuple $M = \langle S, s_0, A, p, r, \gamma \rangle$, where $S$ is a finite set of environment states, $s_0 \in S$ is an initial state, $A$ is a finite set of actions available to the agent, $p : S \times A \to S$ is a transition function, $r : S \times A \times S \to \mathbb{R}$ is a reward function, and $\gamma \in(0, 1]$ is a discount factor. A policy $\pi(a|s)$ is a mapping from the state space $S$ to the action space $A$, that is, $\pi : S \to A$. A trajectory $\tau$ is a sequence of states, actions, and rewards $\langle s_0, a_0, r_1, s_1, a_1, ..., r_T, s_T\rangle$ that describes the agent's interaction with the environment up to a given time step $T$ \cite{skalse2023limitations}. The trajectory return function is
\begin{equation}
 g(\tau)=\sum^{T-1}_{t = 0}\gamma^t r_{t+1}.
 \label{eq:tr-ret}
\end{equation}

At state $s_t$, the agent executes action $a_t$ according to the probability distribution $\pi(a_t|s_t)$. The result of action $a_t$ is the transition to state $s_{t+1}$ according to the probability distribution $p(s_{t+1}|s_t,a_t)$. After transitioning to $s_{t+1}$, the agent receives reward $r_{t+1}$. The process repeats either until episode termination or until reaching a terminal state. The objective is to choose a trajectory $\tau$ such that the trajectory return $g(\tau)$ is maximised. For this, the agent needs to discover an optimal policy $\pi^*$ from interactions with the environment. The optimal policy $\pi^*(a|s_t = s)$ for all $s \in S$ corresponds to the maximum expected return $\mathbb{E}_{\pi^*}[ \, \sum^{T-1}_{k = 0}\gamma^k r_{t+k+1}|s_t = s ]$.

The $Q$-function $q^\pi(s,a)$ in RL quantifies the expected return the agent can achieve by taking a specific action $a$ in a given state $s$ and following a policy $\pi$ thereafter. Formally,
\begin{equation}
q^\pi(s,a) = \mathbb{E}_{\pi}[ \, \sum^{T-1}_{k = 0}\gamma^k r_{t+k+1}|s_t = s, a_t = a ]. \,
\end{equation}
For an optimal policy $\pi^*$, $q^* = q^{\pi^*}$. Every optimal policy $\pi^*$ satisfies the Bellman optimality equation
\begin{equation}
q^*(s, a) = \sum_{s' \in S} p(s'|s, a) (r(s, a, s') + \gamma \max_{a' \in A} q^*(s', a'))
\label{eq:bell}
\end{equation}
for all $a \in A$ and $s \in S$. The policy is optimal if the agent selects an action $a$ greedily  with respect to $q^*(s, a)$. Therefore, the RL task could be solved via solving Eq.~\ref{eq:bell} if we knew the transition probability $p$ for every state of the environment.

Tabular $Q$-learning is an algorithm that does not need knowledge of the transition probability $p$ for estimating optimal $q^*(s, a)$ for all $a \in A$ and $s \in S$ \cite{watkins1992q}. The algorithm estimates $Q$-values from interacting with the environment. The  estimated values $Q(s, a)$ are guaranteed to converge to $q^*(s, a)$ if the agent visits all environment states $s \in S$ infinitely often and takes all possible actions from $s$ infinitely many times.

First, a $Q$-table is initialised randomly for each $s \in S$ and $a \in A$. The $Q(s, a)$ values are then updated at each iteration $i$:
\begin{equation}
 Q_{i+1}(s, a) \xleftarrow{\alpha} r(s, a, s') + \gamma \max_{a'} Q_{i}(s',a'),
\end{equation}
where $\alpha$ is a learning rate. The notation $x \xleftarrow{\alpha} y$ is expanded as $x \xleftarrow{\alpha} x + \alpha(y - x)$.

 Tabular $Q$-learning is a classic RL algorithm known for its simplicity and effectiveness in solving problems with discrete state and action spaces with relatively low dimensionality. This is also the case for the problem used in this work (see Sect.~\ref{section:methods}). To handle more complex problems with continuous or high-dimensional state spaces, deep-learning-based variants have been developed  \cite{sutton2018reinforcement}.

\subsection{Reward Machines} \label{sect:rms}

An RM is a finite-state machine that encapsulates an abstract description of the environment. An RM specifies the reward the agent gets when it transitions between two abstract states in the RM (possibly via a self-loop).
Owing to their state-based design, RMs allow to encode non-Markovian rewards.

Formally, an RM is a tuple $R_\psa = \langle U, u_0, F, \delta_u, \delta_r \rangle$ given a set of propositional symbols $\mathcal P$, a set of environment states $S$, and a set of actions $A$. In the tuple, $U$ is a finite set of states, $u_0$ is an initial state, $F$ is a finite set of terminal states, $\delta_u$ is a state-transition function such that $\delta_u:U \times 2^\props \to U \cup F$, and $\delta_r$ is a state-reward function such that $\delta_r:U \to [S \times A \times S \to \mathbb{R}]$.

At each time step $t$, given an environment experience $\langle s, a, s' \rangle$, a labeling function $L:S \times A \times S \to 2^\props$ assigns truth values to propositions. At each step, the set of propositions that are true in the current environment state is sent to the RM. The state-transition function then decides which abstract successor state is reached. The reward function to be used for the underlying MDP is chosen by the state-reward function.

If it is sufficient to return a real-valued reward rather than a reward function, the RM definition can be simplified. A simple RM is a tuple $R_{\props} = \langle U, u_0, F, \delta_u, \delta_r \rangle$ given a set of propositional symbols $\props$. In the tuple, $U$, $u_0$, $F$, and $\delta_u$ are defined similarly as in $R_\psa$. However, the state-reward function is now $\delta_r:U \times 2^\props \to \mathbb{R}$. In this work, we focus on simple RMs.

Intuitively, an MDP with an RM (MDPRM) is an MDP defined over the cross-product $S \times U$. That is, an MDPRM is a tuple $M_{R} = \langle \tilde{S}, \tilde{s}_0, \tilde{A}, \tilde{p}, \tilde{r}, \tilde{\gamma} \rangle$, where $\tilde{S}$ is a set of states $S \times U$, $\tilde{s}_0 \in \tilde{S}$ is an initial state, $\tilde{A} = A$ is a set of actions available to the agent,
$\tilde{p}(\langle s', u'\rangle | \langle s, u\rangle, a)$
is a state-transition function that now depends on both $u$ and $s$, $\tilde{r}(\langle s, u\rangle, a, \langle s', u'\rangle) = \delta_r(u, L(s, a, s'))$ is a state-reward function,
and $\tilde{\gamma} = \gamma$ is a discount factor. The task formulation with respect to MDPRM is Markovian. Consequently, tabular $Q$-learning (among other RL methods) can be used to solve it. Optimal-solution guarantees of an RL algorithm for MDPRMs are the same as for regular MDPs \cite{icarte2022reward}.

In their work, \inlinecite{icarte2022reward} provided three algorithms for $Q$-learning for an MDPRM. The first algorithm is a baseline for solving MDPRM using tabular $Q$-learning (QRM). It keeps track of the current RM state $u$ and learns Q-values $Q(\langle s, u \rangle, a)$ over the cross-product of the environment and RM states.
As this algorithm does not use the reward structure encoded in an RM, \citeauthor{icarte2022reward} also proposed $Q$-learning with counterfactual experiences (CRM) and hierarchical RL for RMs (HRM).

In CRM, learning happens in the same way as in the baseline algorithm, but counterfactual reasoning is used to generate multiple experiences for a single environment interaction.
Specifically, the reward acquisition at any abstract state $\overline{u} \in U$ for a fixed environment state $s$ is decoupled from the direct interactions with the environment. Therefore, the agent with experience $\langle s, a, s' \rangle$ can simulate synthetic experiences for each RM state without visiting it. These synthetic experiences are called \emph{counterfactual} experiences. Their use in learning makes the algorithms more sample-efficient.

In HRM, the problem is decomposed into sub-problems called \emph{options}, each representing transitions between RM states.
Each option $Y(u, u_t)$ corresponds to the transition in the RM between states $u$ and $u_t$.
The agent learns two policies.
A high-level policy learns to select among the available options to collect the highest reward.
It does so by learning Q-values $Q^H(s, u, u_t)$ that select a state $u_t$ to reach from a given state $u$.
A low-level policy based on $Q_{u,u_t}(s, a)$ learns to act within an option $Y(u,u_t)$.
Noteworthy, synthetic experiences are used for learning $Q_{u,u_t}(s, a)$ for each option.

\section{Numeric Reward Machines}
\label{section:methods}

In this section, we present three RMs that progress from Boolean features to numeric features. The first RM, \bool{}, is the original RM that uses Boolean features. The second RM, \numericbool{}, emulates numeric features using Boolean features. The third RM, \numeric{}, uses numeric features directly. Finally, we analyse and compare the theoretical guarantees of $Q$-learning-based methods with the three introduced RMs with regards to the path lengths they find in the Craft domain. We begin by describing the Craft domain \cite{andreas2017modular}, which serves as a running example and is the basis for our experimental comparisons.

\subsection{Environment}

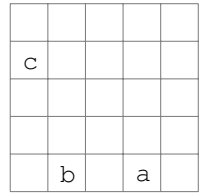
\begin{figure}[tbp]
\centering
\begin{tikzpicture}
\draw[step=0.5,gray,thin] (0,0) grid (2.5,2.5);
\draw (2,0.21) node[anchor=east] {\texttt{a}};
\draw (0.5,1.71) node[anchor=east] {\texttt{c}};
\draw (1,0.24) node[anchor=east] {\texttt{b}};
\end{tikzpicture}
\caption{Example grid map inspired by the Craft domain \cite{andreas2017modular}. It contains objects \texttt{a}, \texttt{b}, and \texttt{c}. }
\label{fig:5-5-grid}
\end{figure}

We use environments inspired by the Craft domain \cite{andreas2017modular}. It is a finite-grid world with objects of different types. The RL agent is instructed to visit some objects. The world contains no obstacles besides the surrounding walls, and its dimensions are fixed. We show an example environment with three objects in Fig.~\ref{fig:5-5-grid}.

\subsection{\bool{} Reward Machines}

\begin{figure}[tbp]
\centering
\begin{tikzpicture}[node distance = 2cm, on grid, auto]

    \node (u0) [state, initial text = {}] {$u_a$};
    \node (u1) [state, right = of u0] {$u_b$};
    \node (u2) [state, right = of u1] {$u_c$};
    \node (u3) [state, accepting, right = of u2] {$u$};

    \path [-stealth, thick]
        (u0) edge node[above right, pos = 0]  {$\langle$\texttt{a}; $0$$\rangle$}   (u1)
        (u0) edge [loop above] node {$\langle$$\lnot$ \texttt{a}; $0$$\rangle$}   (u0)
        (u1) edge node[above right, pos = 0]  {$\langle$\texttt{b}; $0$$\rangle$}   (u2)
        (u1) edge [loop above] node {$\langle$$\lnot$ \texttt{b}; $0$$\rangle$}   (u1)
        (u2) edge node[above right, pos = 0]  {$\langle$\texttt{c}; $1$$\rangle$}   (u3)
        (u2) edge [loop above] node {$\langle$$\lnot$ \texttt{c}; $0$$\rangle$}   (u2);

\end{tikzpicture}
\caption{\bool{} reward machine with Boolean features \texttt{a}, \texttt{b}, and \texttt{c} for a sequential task \texttt{a-b-c}. }
\label{fig:rm-abc-bool}
\end{figure}
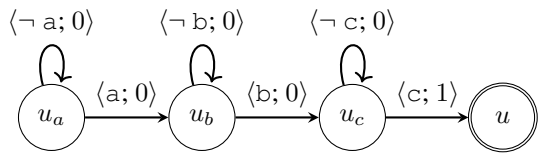

The concept of \bool{} RMs was introduced by \citet{icarte2022reward}. In \bool{} RMs, the features are propositions $p \in \props$ that can become true in the environment given experience $\langle s, a, s' \rangle$.
Figure~\ref{fig:rm-abc-bool} shows an example of such a \bool{} RM for a sequential task \texttt{a-b-c}, where the agent has to visit three objects of types \texttt{a}, \texttt{b}, and \texttt{c}, in order. The agent receives reward $r$ after taking a transition labeled with
$\langle \{ P \}; r \rangle$ in the depicted RM, where a set of propositions $P \subseteq \props$ is currently true in the environment.

To address the common problem of reward sparsity, one typically uses potential-based reward shaping. \citet{ng1999policy} proved that, for any given MDP $\langle S, s_0, A, r, p, \gamma \rangle$ and a potential function $\Phi : S \to \mathbb{R}$, the reward function $r$ can be changed to $r'(s, a, s') = r(s, a, s') + \gamma \Phi(s') - \Phi(s)$ without changing the set of optimal policies. Like \citet{icarte2022reward}, we compute the potential function $\Phi$ using value iteration over RM states, treating the RM as an MDP.

The idea of automatic reward shaping is to change the reward function such that the policy is easier to learn. To this end, intermediate rewards are given to the agent as it approaches the target. However, these intermediate rewards do not carry any interpretable meaning. They are only tweaked for finding the optimal policy fast. To obtain a method that preserves the meaning of rewards, we propose to use numeric features to shape the reward function.

\subsection{\Numericbool{} Reward Machines}

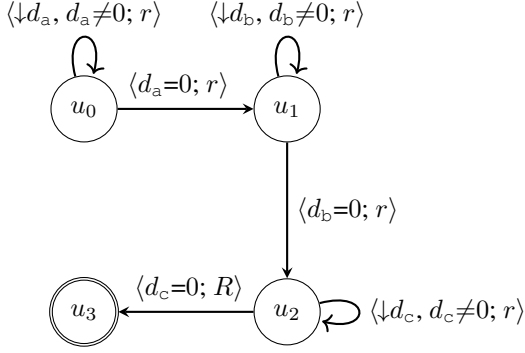
\begin{figure}[tbp]
\centering
\begin{tikzpicture}[node distance = 2.7cm, on grid, auto]

    \node (u0) [state, initial text = {}] {$u_0$};
    \node (u1) [state, right = of u0] {$u_1$};
    \node (u2) [state, below = of u1] {$u_2$};
    \node (u3) [state, accepting, left = of u2] {$u_3$};

    \path [-stealth, thick]
        (u0) edge node[above right, pos = 0]  {$\langle$$d_\texttt{a}$=$0$; $r$$\rangle$}   (u1)
        (u0) edge [loop above] node {$\langle$$\dec d_\texttt{a}$, $d_\texttt{a}$$\neq$$0$; $r$$\rangle$}   (u0)
        (u1) edge node[right]  {$\langle$$d_\texttt{b}$=$0$; $r$$\rangle$}   (u2)
        (u1) edge [loop above] node {$\langle$$\dec d_\texttt{b}$, $d_\texttt{b}$$\neq$$0$; $r$$\rangle$}   (u1)
        (u2) edge node[above left, pos = 0]  {$\langle$$d_\texttt{c}$=$0$; $R$$\rangle$}   (u3)
        (u2) edge [loop right] node {$\langle$$\dec d_\texttt{c}$, $d_\texttt{c}$$\neq$$0$; $r$$\rangle$}   (u2);

\end{tikzpicture}
\caption{\Numericbool{} reward machine with Boolean features $\dec d_\texttt{a}$, $\dec d_\texttt{b}$, $\dec d_\texttt{c}$, $d_\texttt{a}$=$0$, $d_\texttt{b}$=$0$, and $d_\texttt{c}$=$0$ for a sequential task \texttt{a-b-c}.
}
\label{fig:rm-abc-num-bool}
\end{figure}

The first type of RM that we introduce in this paper is a \numericbool{} RM. This concept builds directly on \bool{} RMs. Instead of automatic reward shaping, we introduce a numeric feature  $d_\texttt{a}$---distance between agent \texttt{A} and target \texttt{a}. We translate $d_\texttt{a}$ to two Boolean features $\dec d_\texttt{a}$ and $d_\texttt{a}$=$0$. Feature $\dec d_\texttt{a}$ becomes true if the distance between the agent and object \texttt{a} decreases. Feature $d_\texttt{a}$=$0$ becomes true when the agent reaches object \texttt{a}.
The agent is rewarded with a fixed reward $r > 0$ when $\dec d_\texttt{a}$ becomes true. When $d_\texttt{a}$=$0$ becomes true, the agent is rewarded with $R > 0$. An example RM is shown in Fig.~\ref{fig:rm-abc-num-bool}.

In a \bool{} RM with automatic reward shaping, the agent is rewarded for each experience $\langle s, a, s' \rangle$ by an automatically generated positive reward \cite{icarte2022reward}. In contrast, in a \numericbool{} RM, the agent is rewarded
by a constant reward $r$ or $R$ for a part of experiences $\langle s, a, s' \rangle$ when $\dec d_\texttt{a}$ or $d_\texttt{a}$=$0$ becomes true, respectively. In all other cases, the agent receives a reward of $0$. In this way, the agent is positively reinforced to approach the target. Exposing the \numericbool{}-RM structure to the agent in CRM and HRM has the potential to boost learning similar to the original approach.

Ideally, the agent should receive a reward that depends on $d_\texttt{a}$. In this way, the agent will be meaningfully guided by each experience $\langle s, a, s' \rangle$. To realise this, we introduce \numeric{} RMs.

\subsection{\Numeric{} Reward Machines}

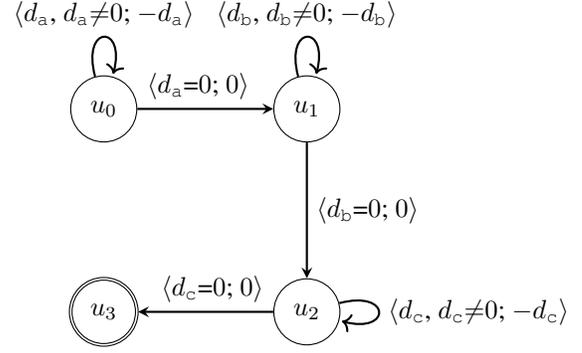
\begin{figure}[tbp]
\centering
\begin{tikzpicture}[node distance = 2.7cm, on grid, auto]

    \node (u0) [state, initial text = {}] {$u_0$};
    \node (u1) [state, right = of u0] {$u_1$};
    \node (u2) [state, below = of u1] {$u_2$};
    \node (u3) [state, accepting, left = of u2] {$u_3$};

    \path [-stealth, thick]
        (u0) edge node[above right, pos = 0]  {$\langle$$d_\texttt{a}$=$0$; $0$$\rangle$}   (u1)
        (u0) edge [loop above] node {$\langle$$d_\texttt{a}$, $d_\texttt{a}$$\neq$$0$; $-d_\texttt{a}$$\rangle$}   (u0)
        (u1) edge node[right]  {$\langle$$d_\texttt{b}$=$0$; $0$$\rangle$}   (u2)
        (u1) edge [loop above] node {$\langle$$d_\texttt{b}$, $d_\texttt{b}$$\neq$$0$; $-d_\texttt{b}$$\rangle$}   (u1)
        (u2) edge node[above left, pos = 0]  {$\langle$$d_\texttt{c}$=$0$; $0$$\rangle$}   (u3)
        (u2) edge [loop right] node {$\langle$$d_\texttt{c}$, $d_\texttt{c}$$\neq$$0$; $-d_\texttt{c}$$\rangle$}   (u2);

\end{tikzpicture}
\caption{\Numeric{} reward machine with numeric features $d_\texttt{a}$, $d_\texttt{b}$, and $d_\texttt{c}$ and Boolean features $d_\texttt{a}$=$0$, $d_\texttt{b}$=$0$, and $d_\texttt{c}$=$0$ for a sequential task \texttt{a-b-c}. }
\label{fig:rm-abc-num}
\end{figure}

The second new type of RM is a \numeric{} RM. Here, we use the numeric feature $d_\texttt{a}$ along with the Boolean feature $d_\texttt{a}$=$0$ directly in the RM. In this RM, the Boolean feature $d_\texttt{a}$=$0$ governs the transition between the RM states, and the numeric feature $d_\texttt{a}$ is used for rewarding the agent with $-d_\texttt{a}$. In this way, the agent is rewarded for each experience by the negative distance to the target. The reward becomes $0$ upon arrival at the target. Rewarding the agent is therefore controlled by the real conditions in the environment. An example \numeric{} RM is shown in Fig.~\ref{fig:rm-abc-num}.

Noteworthy, as the RM structure is still exposed to the agent via the Boolean feature $d_\texttt{a}$=$0$, learning can be boosted by CRM or HRM.

\subsection{Shortest-Path Guarantees}
\label{sect:shortest-path-solution-guarantees}

\begin{figure}[tbp]
    \centering
    \begin{tikzpicture}
    \draw[step=0.5,gray,thin] (0,0) grid (2,2);
    \draw (1,1.23) node[anchor=east] {\texttt{A}};
    \draw (2.1,0.23) node[anchor=east] {\texttt{a}$_1$};
    \draw (0.6,1.73) node[anchor=east] {\texttt{a}$_2$};
    \end{tikzpicture}
    \caption{Agent \texttt{A} can choose to approach either \texttt{a}$_1$ or \texttt{a}$_2$.}
    \label{fig:task-1-no-search-sol-gur}
\end{figure}
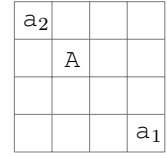

Next, we investigate whether the discovered policies by a $Q$-learning-based method correspond to the shortest paths between the agent and the targets for the proposed RMs. Ideally, we would like our RMs to have this property to avoid tweaking the reward values in the RM for each new task.
For this analysis, we use an example map, shown in Fig.~\ref{fig:task-1-no-search-sol-gur}, with two targets \texttt{a}$_1$ and \texttt{a}$_2$, both of type \texttt{a}.
The agent is instructed to visit one of the targets and can freely choose whether to approach \texttt{a}$_1$ or \texttt{a}$_2$.
We choose this example as an extreme case, where the agent approaches exactly one of the targets with every move.
This can ``confuse'' the agent, as it receives a positive reward for any move, and thus might lead to finding sub-optimal paths.
In addition, we consider a simplified scenario with only one target \texttt{a} on the map (at any location).

Since the agent performs one move action at each time step, the path that it takes to either target is optimal if the number of episode steps $T$ is minimum. To check whether this is the case, we dissect the trajectory return functions (Eq.~\ref{eq:tr-ret}) with respect to $T$ for the introduced RMs. In all our calculations, we assume $\gamma < 1$.

\subsubsection{\bool{} Reward Machines}
The agent receives reward $R$ when it arrives at a terminal state and zero otherwise, i.e.,
\begin{equation}
    r_t =
    \begin{cases}
        R,& t = T\\
        0,& t \neq T.
    \end{cases}
\end{equation}
Therefore, the return for any trajectory is
\begin{equation}
 g(T) = \gamma^{T-1} R.
\end{equation}
As the agent seeks to maximise the accumulated reward, the lower the episode length $T$, the higher the return $g(T)$. Thus, \bool{} RMs guarantee to find shortest paths.
This holds for both the example map with two targets and the simplified scenario with one target.

\subsubsection{\Numericbool{} Reward Machines}

A reward $r > 0$ is given for decreasing the distance towards any target. Consequently, for the example map in Fig.~\ref{fig:task-1-no-search-sol-gur}, where the targets are in the corners of the map, the agent receives a positive reward for any move, i.e.,
\begin{equation}
    r_t =
    \begin{cases}
        R,& t = T\\
        r,& t \neq T.
    \end{cases}
\end{equation}

The return for any trajectory is $g(T) = \gamma^{T-1}R + r \sum^{T-2}_{t = 0} \gamma^t$. The geometric-series solution \cite{zorich2016mathematical} yields
\begin{equation}
    g(T) = \gamma^{T-1}R + r \frac{1-\gamma^{T-1}}{1 - \gamma}.
\end{equation}
The second term in this equation increases with the number of steps $T$. Therefore, the resulting path length might not be minimum.

Assume that a shortest path uses $T^*$ steps, resulting in the trajectory return
\begin{equation}
    g(T^*) = \gamma^{T^*-1}R + r \frac{1-\gamma^{T^*-1}}{1 - \gamma}.
    \label{eq:G-0-task-2-opt}
\end{equation}
If the agent takes two additional back-and-forth steps $n$ times, the corresponding return is
\begin{equation}
    g(T^*+2n) = \gamma^{T^*-1+2n}R + r \frac{1-\gamma^{T^*-1+2n}}{1 - \gamma}.
    \label{eq:G-0-task-2-subopt}
\end{equation}
If $g(T^*)$ is larger than $g(T^*+2n)$, the resulting path is guaranteed to have the shortest length. The comparison yields
\begin{align}
\begin{split}
    \gamma^{T^*-1}R + r \frac{1-\gamma^{T^*-1}}{1 - \gamma} > \gamma^{T^*-1+2n}R + r \frac{1-\gamma^{T^*-1+2n}}{1 - \gamma}; \\
    \gamma^{T^*-1}R(1-\gamma^{2n}) > \frac{r\gamma^{T^*-1}}{1 - \gamma}(1-\gamma^{2n}); \\
    r < R(1-\gamma).
\end{split}
\end{align}

Therefore, we find a shortest path if $r$ is lower than $R(1-\gamma)$. For example, if $R = 1$ and $\gamma = 0.9$, then $r$ needs to be lower than $0.1$.

Next, we consider the simplified scenario with one target \texttt{a}.
The agent gets a positive reward $r$ if the distance to \texttt{a} decreases.
The distance from the agent to the target at step $t$ is written as $d_t$, and
the change in distance from step $t$ to $t+1$ is captured by $\Delta d = d_{t+1} - d_t$.
If $\Delta d < 0$, then the agent approaches the target and receives reward $r$, i.e.,
\begin{equation}
    r_t =
    \begin{cases}
        R,& \Delta d = 0,\\
        r,& \Delta d < 0, \\
        0,& \Delta d > 0.
    \end{cases}
\end{equation}
Now, if the agent takes $2n$ additional back-and-forth steps, the trajectory return is
\begin{align}
\begin{split}
    g_o(T^*+2n) = \gamma^{T^*-1+2n}R + r \frac{1-\gamma^{T^*-1}}{1 - \gamma} +\\
    +r\gamma^{T^*-1}(\gamma + \gamma^3 + ... + \gamma^{2n-1}),
\end{split}
\end{align}
where the subscript $o$ denotes that only one \texttt{a} is on the map.
The final trajectory return is then rewritten using the geometric-series solution as follows:
\begin{align}
\begin{split}
    g_o(T^*+2n) = \gamma^{T^*-1+2n}R + r \frac{1-\gamma^{T^*-1}}{1 - \gamma} +\\
    +r\gamma^{T^*-1}\frac{\gamma-\gamma^{2n+1}}{1-\gamma^2}.
\end{split}
\end{align}

The comparison between $g_o(T^*)$ and $g_o(T^*+2n)$ yields
\begin{align}
\begin{split}
    \gamma^{T^*-1}R  >
     \gamma^{T^*-1+2n}R +r\gamma^{T^*-1}\frac{\gamma-\gamma^{2n+1}}{1-\gamma^2}; \\
    r < R\frac{1-\gamma^2}{\gamma}.
\end{split}
\end{align}
Here, $r$ needs to be smaller than $R\frac{1-\gamma^2}{\gamma}$ to guarantee that we find a shortest path.
For example, if $R = 1$ and $\gamma=0.9$, then $r$ needs to be smaller than $0.21$.

We showed that rewards in \numericbool{} RMs should be selected carefully. This solution dependence on the reward values is problematic and could be addressed in future work.

\subsubsection{\Numeric{} Reward Machines}

For several targets on the map, the reward scheme of \numeric{} RMs can be as follows: The agent is rewarded with $-d$, where $d$ is the lowest distance to any target of the same type. We hypothesise that this modification preserves the guarantee to find shortest paths. However, we leave the proof of this conjecture for future work.

For the case with a single target \texttt{a}, \numeric{} RMs reward the agent with
\begin{equation}
    r_t =-d,
\end{equation}
where $d$ is the number of steps to the closest target. The shortest-trajectory return is then
\begin{equation}
    g_o(T^*) = \sum^{T^*-1}_{t = 0}\gamma^t(t-T^*).
\end{equation}
If the agent takes $2n$ additional back-and-forth steps just before reaching the target, the trajectory return is
\begin{align}
\begin{split}
    g_o(T^*+2n) = \sum^{T^*-1}_{t = 0}\gamma^t(t-T^*)-\\
    -2\gamma^{T^*-1}\sum^n_{k = 0}\gamma^{2k-2}-1\gamma^{T^*-1}\sum^{n-1}_{k=0}\gamma^{2k+1},
\end{split}
\end{align}
where the last two series correspond to the rewards received at step ``back'' and ``forth'', respectively. The difference between $g_o(T^*)$ and $g_o(T^*+2n)$ will always be positive because the last two terms in $g_o(T^*+2n)$ will always be negative.
Therefore, \numeric{} RMs guarantee shortest-path solutions regardless of the selected parameters for a map with one target.
In future work, we plan to extend these proofs to sequential tasks, where the agent needs to visit multiple targets.

\section{Experimental Evaluation}

\begin{figure}[tbp]
\centering
\includegraphics[width=0.48\textwidth]{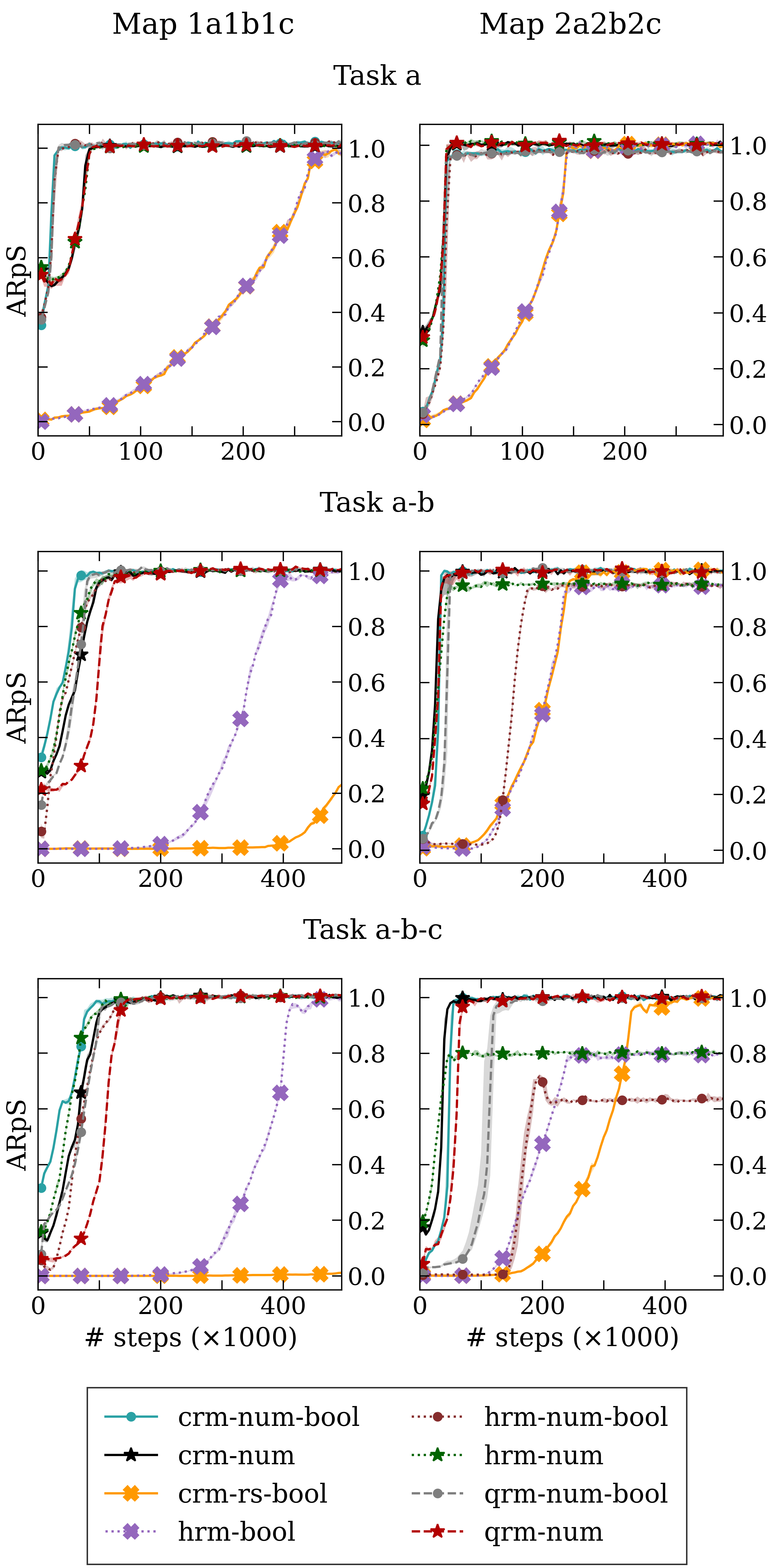}
\caption{Median performance for map \texttt{1a1b1c} (left column) and map \texttt{2a2b2c} (right column). The $25$th and $75$th percentiles are shown in the shadowed regions. ARpS stands for average reward per step.}
\label{fig:results}
\end{figure}

For our experimental evaluation, we choose a $41\times41$ grid, matching the grid size in the Craft domain tested by \citet{icarte2022reward}.
We use two different object setups: ($1$) \texttt{1a1b1c} and ($2$) \texttt{2a2b2c}.
Letters (\texttt{a}, \texttt{b}, and \texttt{c}) encode the object type, and the numeral in front of each letter (\texttt{1} and \texttt{2}) encodes how often the object type appears on the map.
That is, in setup \texttt{1a1b1c}, three objects, \texttt{a}, \texttt{b}, and \texttt{c}, are randomly placed on the grid. Likewise, in setup \texttt{2a2b2c}, two pairs of objects \texttt{a}, \texttt{b}, and \texttt{c} are randomly placed on the grid. In the latter setup, the agent is expected to discover which objects of the same type are closest to its initial location. We use Manhattan distance as the distance metric in this work, i.e., the grid cells are four-connected. The initial agent position is fixed throughout the experiments.
For each setup, we generate $10$ random maps to check the variability of the results.

For a given map, the agent is instructed to visit objects sequentially in three different tasks: \texttt{a}, \texttt{a-b}, and \texttt{a-b-c}. In task \texttt{a}, the agent needs to visit only an object of type \texttt{a}. In task \texttt{a-b}, the agent needs to visit object types \texttt{a} and then \texttt{b}. Finally, in task \texttt{a-b-c}, the agent needs to visit object types \texttt{a}, \texttt{b}, and \texttt{c}, in order. We show results only for two maps (stemming from setups \texttt{1a1b1c} and \texttt{2a2b2c}) because the relative algorithm performance is similar on all generated maps of one setup. One of these maps, with setup \texttt{2a2b2c}, is provided in the appendix in Fig.~\ref{fig:real-map} for discussion purposes. The median results for all $10$ maps are shown in the appendix in Fig.~\ref{fig:results-all}.

For each map, we run each algorithm six times and measure the average reward per step. All rewards are normalised to be between $0$ and $1$. A reward of $1$ corresponds to the optimal policy found via value iteration \cite{icarte2022reward}. We report the median performance over the six runs and the $25$th and $75$th percentiles in Fig.~\ref{fig:results} (barely visible for most algorithms).

We compare the following combinations of methods. First, the best-performing methods for Boolean features and tabular domains are CRM with reward shaping (crm-rs-bool) and HRM (hrm-bool) according to \citet{icarte2022reward}. Despite being faster than crm-rs-bool, hrm-bool tends to converge to sub-optimal policies. We use both methods as baselines in our experiments.
For \numericbool{} and \numeric{} RMs, we use CRM, HRM, and $Q$-learning, abbreviated as crm-num-bool, hrm-num-bool, qrm-num-bool, crm-num, hrm-num, and qrm-num. The reason for including $Q$-learning together with \numericbool{} and \numeric{} RMs is to explore whether the knowledge of the RM structure boosts the agent's learning in CRM and HRM. The parameters in the $Q$-learning, CRM, and HRM methods are taken from \citet{icarte2022reward}.

\section{Results and Discussion}

Using \bool{} RMs, the agent receives a reward of $1$ after transitioning to the terminal state and a reward of $0$ for all other RM transitions. Regardless of the number of targets, the agent solves all evaluated tasks with the shortest path.

In \numericbool{} RMs, we choose $r = 0.1$ and $R = 1000$ to guarantee that the agent finds shortest paths (see Sect.~\ref{sect:shortest-path-solution-guarantees}). Indeed, the agent finds shortest paths in all tasks.

With \numeric{} RMs, the agent finds a shortest path only for task \texttt{a}, for which we proved that the optimal policy corresponds to the shortest-path solution in Sect.~\ref{sect:shortest-path-solution-guarantees}. However, for tasks \texttt{a-b} and \texttt{a-b-c}, the agent converges to non-optimal solutions. It is only able to find a shortest path once we change the terminal rewards from $0$ to values of $10\,000$ and $100\,000$, respectively. We will investigate this outcome in future work.

We visualise the results in Fig.~\ref{fig:results}.
The shadowed regions depicting the $25$th and $75$th percentiles are narrow. This is because randomisation in our experiments comes only from tie-breaking in the case of equal $Q$-values and from using $\epsilon$-greedy for exploration.
The methods using \bool{} RMs converge slower than other methods.

In tasks \texttt{a-b} and \texttt{a-b-c}, all HRM-based methods converge to sub-optimal policies on map \texttt{2a2b2c}. This can be explained as follows. As shown in Fig.~\ref{fig:real-map}, the Manhattan distances to the top-positioned \texttt{a} and bottom-positioned \texttt{a} from the initial position of agent \texttt{A} are $20$ and $21$, respectively. The HRM-based agent chooses the top-positioned \texttt{a} because it is always greedy towards the current transition in the RM. The total distances to the top-positioned \texttt{a-b} and bottom-positioned \texttt{a-b} are $27$ and $26$, respectively. Consequently, after choosing the top route, the HRM-based algorithms converge to a sub-optimal policy. In task \texttt{a-b-c}, the performance of the HRM-based algorithms deteriorates further. This is explained by the total distances to the top-positioned \texttt{a-b-c} and bottom-positioned \texttt{a-b-c} being $40$ and $32$, respectively.

Furthermore, the results show that the CRM-based methods always converge at least as fast as pure-$Q$-learning-based methods. Indeed, the knowledge of the RM structure speeds up learning.

Another observation is that all CRM- and pure-$Q$-learning-based methods converge faster on map type \texttt{2a2b2c} than on map type \texttt{1a1b1c}. This is because placing more objects of the same kind on the map makes it more likely that the total distance to the final object is shorter from the initial position of the agent.

On map type \texttt{1a1b1c}, the \numericbool{}-RM-based methods outperform the corresponding \numeric{}-RM-based methods. In contrast, on map type \texttt{2a2b2c}, the \numeric{}-RM-based methods perform at least as well as the corresponding \numericbool{}-RM-based methods. We believe that this result is related to the numeric reward scheme.
Furthermore, the \numericbool{}-RM-based methods converge to a value slightly below $1.0$ in task \texttt{a} on map type \texttt{2a2b2c}.
We will investigate both of these phenomena in the future.

The performance difference between the methods increases with difficulty from task \texttt{a} to task \texttt{a-b-c}. In task \texttt{a}, all methods from the same feature category perform similarly. That is, crm-rs-bool and hrm-bool perform similar; qrm-num-bool, crm-num-bool and hrm-num-bool perform similar; and finally, qrm-num, qrm-num-bool, and hrm-num perform similar. The explanation is the task simplicity. Only once the task becomes challenging enough, performance differences start to appear.

Overall, the results show that \numericbool{} and \numeric{} RMs speed up learning in comparison with \bool{} RMs with automatic reward shaping.

\section{Conclusions}

In this study, we extend RMs with numeric features by introducing \numericbool{} and \numeric{} RMs. We compare them with the original \bool{} RMs introduced by \citet{icarte2022reward}.

For our experimental evaluation, we use the Craft domain. This is a simple grid world without obstacles, allowing the use of tabular RL methods. We test the developed methods in two directions. First, the results confirm that the developed reward acquisition schemes in \numericbool{} and \numeric{} RMs outperform automatic reward shaping used in \bool{} RMs. Second, the results confirm that the exposure of the \numericbool{}- and \numeric{}-RM structures to the agent speeds up learning in CRM and HRM. The advantage of \numeric{} RMs over \numericbool{} RMs is not yet clear. Additional experiments with more complicated tasks should be performed to make meaningful conclusions.

Furthermore, we prove the shortest-path guarantees for the task with one target type. We show that \bool{} RMs guarantee finding shortest paths regardless of the reward values in the RM. In contrast, in \numericbool{} RMs, the terminal reward $R$ should be higher than the intermediate reward $r$, and the difference between $R$ and $r$ will depend on the task. For \numeric{} RMs, we prove that shortest paths are guaranteed for tasks with one target type. However, for sequential tasks with several targets, the experimental results show that this is not the case.

In future work, it will be interesting to test the \numericbool{} and \numeric{} RMs in tabular domains with obstacles, continuous-space domains, and continuous-control tasks. In addition, we will perform experiments for unordered tasks, where the agent can visit the assigned targets in any order. Furthermore, we will extend the proofs of the shortest-path guarantees for sequential tasks with several targets. Ideally, the problem of the solution dependence on the reward values in the RM should be resolved.

\section{Acknowledgments}
This work was supported by Ericsson Research and the Wallenberg AI, Autonomous Systems, and Software Program (WASP) funded by the Knut and Alice Wallenberg Foundation. The computations were enabled by resources provided by the National Academic Infrastructure for Supercomputing in Sweden (NAISS), partially funded by the Swedish Research Council through grant agreement no. $2022$-$06725$.

\bibliography{aaai24}

\renewcommand\thefigure{A.\arabic{figure}}
\section*{Appendix}
\setcounter{figure}{0}

\begin{figure}[h]
\centering
\includegraphics[width=0.48\textwidth]{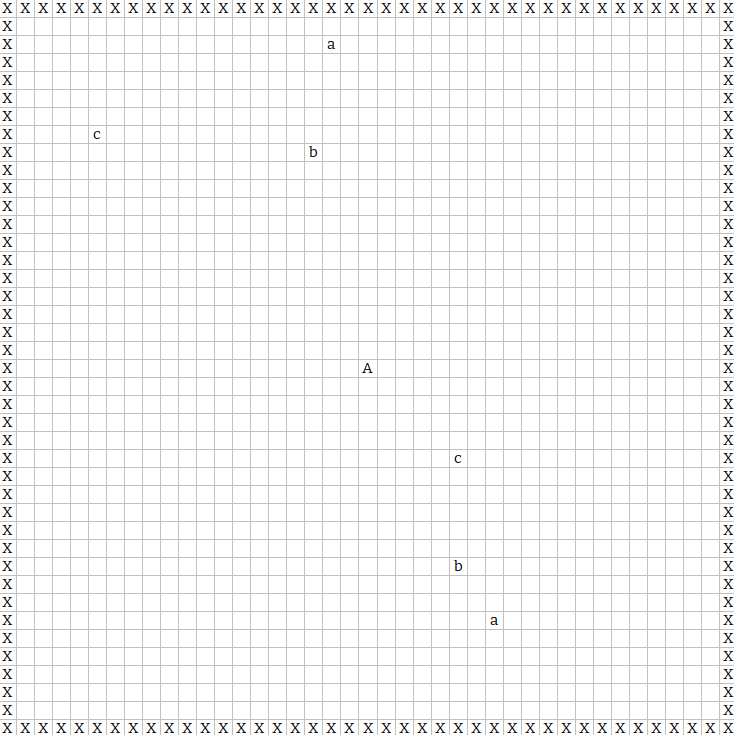}
\caption{Analysed map with setup \texttt{2a2b2c}, agent \texttt{A}, and walls \texttt{X}.}
\label{fig:real-map}
\end{figure}

\begin{figure}[tbp]
\centering
\includegraphics[width=0.48\textwidth]{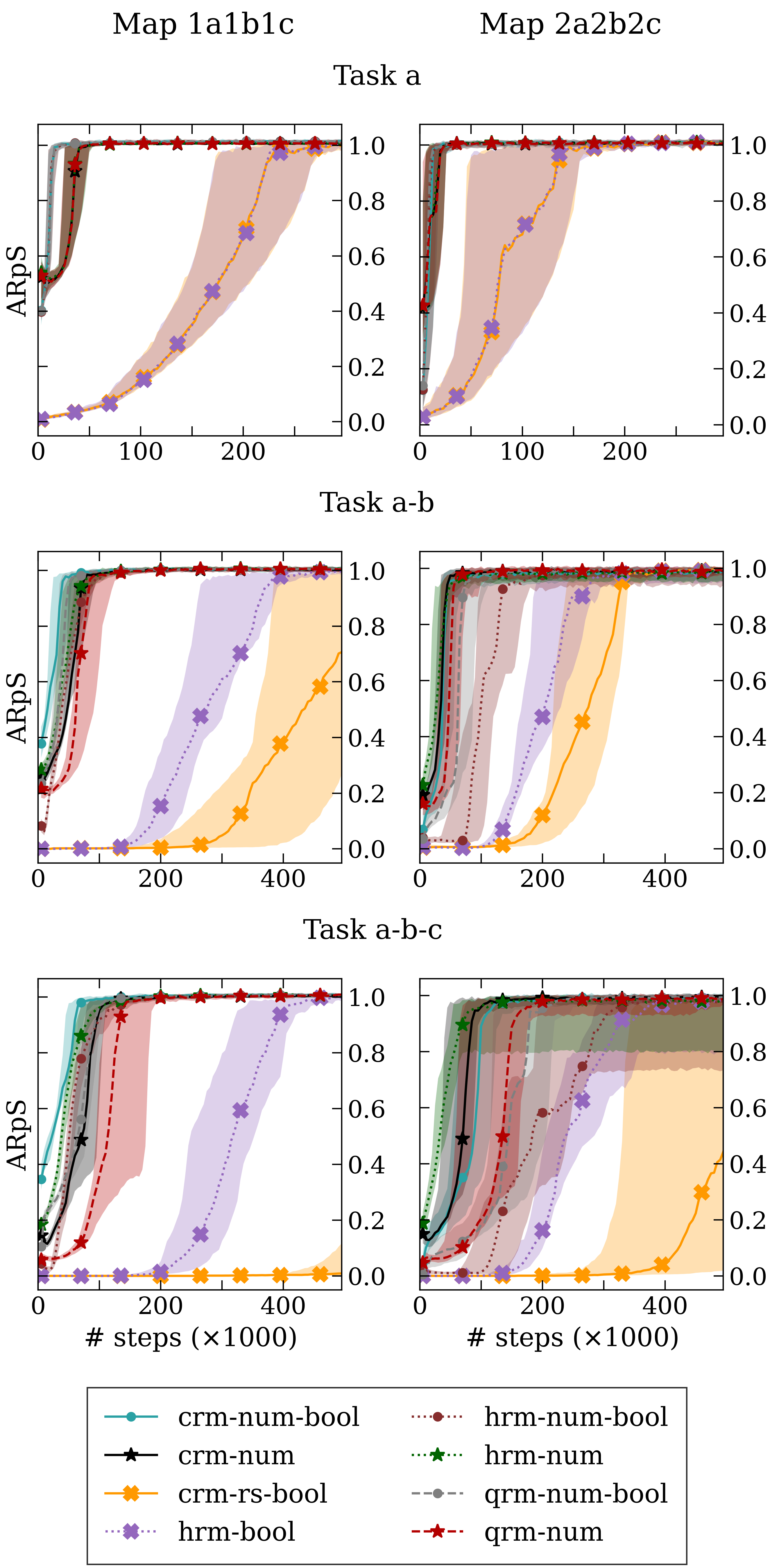}
\caption{Median performance for all $10$ \texttt{1a1b1c} and \texttt{2a2b2c} maps (left and right columns, respectively). The $25$th and $75$th percentiles are shown in the shadowed regions. ARpS stands for average reward per step.}
\label{fig:results-all}
\end{figure}

\end{document}